\def\paperID{0023}
\def\confName{CVPR}
\def\confYear{2024}

\def\paperTitle{In The Wild Ellipse Parameter Estimation for Circular Dining Plates and Bowls}

\def\authorBlock{
    Akil Pathiranage\textsuperscript{1}\thanks{authors contributed equally} , Chris Czarnecki\textsuperscript{1}\textsuperscript{*}, Yuhao Chen\textsuperscript{1}\thanks{corresponding author} , Pengcheng Xi\textsuperscript{2}, Linlin Xu\textsuperscript{1}, Alexander Wong\textsuperscript{1}\\
    \textsuperscript{1} Vision and Image Processing Lab, University of Waterloo,
    \textsuperscript{2} National Research Council Canada\\
    {\tt\small \{alkarago,cczarnec,yuhao.chen1\}@uwaterloo.ca} \\
    {\tt\small pengcheng.xi@nrc-cnrc.gc.ca} \\
    {\tt\small \{l44xu,alexander.wong\}@uwaterloo.ca}

}

\newif\ifreview 
\newif\ifarxiv \newcommand{\arxiv}{\arxivtrue}
\newif\ifcamera 
\newif\ifrebuttal 

\arxiv

\pdfoutput=1
\documentclass[10pt,twocolumn,letterpaper]
{article}
\ifreview \usepackage[review]{cvpr} \fi
\ifarxiv \usepackage[pagenumbers]{cvpr} \fi
\ifrebuttal \usepackage[rebuttal]{cvpr} \fi
\ifcamera \usepackage{cvpr} \fi

\usepackage{graphicx}	
\usepackage{amsmath}	
\usepackage{amssymb}	
\usepackage{booktabs}
\usepackage{times}
\usepackage{microtype}
\usepackage{epsfig}
\usepackage[table,xcdraw,dvipsnames]{xcolor}
\usepackage{caption}
\usepackage{float}
\usepackage{placeins}
\usepackage{color, colortbl}
\usepackage{stfloats}
\usepackage{enumitem}
\usepackage{tabularx}
\usepackage{xstring}
\usepackage{multirow}
\usepackage{xspace}
\usepackage{url}
\usepackage{subcaption}
\usepackage{xcolor}
\usepackage[hang,flushmargin]{footmisc}

\ifcamera \usepackage[accsupp]{axessibility} \fi

% ...

\ifarxiv  \fi

\newcommand{\R}[1]{{%
    \textbf{%
        \ifstrequal{#1}{1}{\textcolor{red}{R#1}}{%
        \ifstrequal{#1}{2}{\textcolor{blue}{R#1}}{%
        \ifstrequal{#1}{3}{\textcolor{magenta}{R#1}}{%
        \ifstrequal{#1}{4}{\textcolor{teal}{R#1}}{%
                           \textcolor{cyan}{R#1}%
        }}}}%
    }%
}}

\usepackage{xr-hyper}

\makeatletter
\newcommand*{\addFileDependency}[1]{
  \typeout{(#1)}
  \@addtofilelist{#1}
  \IfFileExists{#1}{}{\typeout{No file #1.}}
}

\makeatother
\newcommand*{\myexternaldocument}[1]{
    \externaldocument{#1}
    \addFileDependency{#1.tex}
    \addFileDependency{#1.aux}
}

\definecolor{cvprblue}{rgb}{0.21,0.49,0.74}
\usepackage[pagebackref,breaklinks,colorlinks,citecolor=cvprblue]{hyperref}
\usepackage[capitalize]{cleveref}
\crefname{section}{Sec.}{Secs.}
\crefname{table}{Table}{Tables}
\crefname{figure}{Fig.}{Figs.}

\ifarxiv \crefname{appendix}{App.}{Apps.}
\else \crefname{appendix}{Suppl.}{Suppls.} \fi

\usepackage{tikz}
\usepackage{tikzit}
\usepackage{amsmath}
\usepackage{newtxmath,tikz-cd}
\usepackage{xcolor}
\usepackage{pgfplots}
\usetikzlibrary{calc}
\usetikzlibrary{decorations.pathreplacing,fit,shapes.geometric}
\usepackage{mathtools}
\usepackage{adjustbox}
\usepackage{figs/quiver}
\usepackage{afterpage}
\usepackage{placeins}
\usepackage{svg}
\unless\ifarxiv \myexternaldocument{_supplementary} \fi

\tikzstyle{point}=[fill={rgb,255: red,226; green,226; blue,226}, draw={rgb,255: red,111; green,111; blue,111}, shape=circle]

\tikzstyle{ellipse-edge}=[-, draw=black, line width=1.5pt]
\tikzstyle{step-arrow}=[draw={rgb,255: red,198; green,84; blue,49}, ->, line width=1.5pt]

\pgfplotsset{compat=1.18}
\begin{document}
\title{\paperTitle}
\author{\authorBlock}
\maketitle

\begin{abstract}
Ellipse estimation is an important topic in food image processing because it can be leveraged to parameterize plates and bowls, which in turn can be used to estimate camera view angles and food portion sizes.
Automatically detecting the elliptical rim of plates and bowls and estimating their ellipse parameters for data \emph{in-the-wild}  is challenging: diverse camera angles and plate shapes could have been used for capture, noisy background, multiple non-uniform plates and bowls in the image could be present. Recent advancements in foundational models offer promising capabilities for zero-shot semantic understanding and object segmentation. However, the output mask boundaries for plates and bowls generated by these models often lack consistency and precision compared to traditional ellipse fitting methods. In this paper, we combine ellipse fitting with semantic information extracted by zero-shot foundational models and propose WildEllipseFit, a method to detect and estimate the elliptical rim for plate and bowl. Evaluation on the proposed Yummly-ellipse dataset demonstrates its efficacy and zero-shot capability in real-world scenarios.
\end{abstract}
\begin{figure}[h]
    \centering
        \subfloat[]{{\includegraphics[width=.24\linewidth]{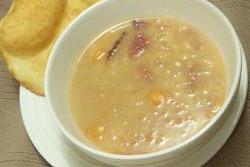} }}%
        \subfloat[]{{\includegraphics[width=.24\linewidth]{ 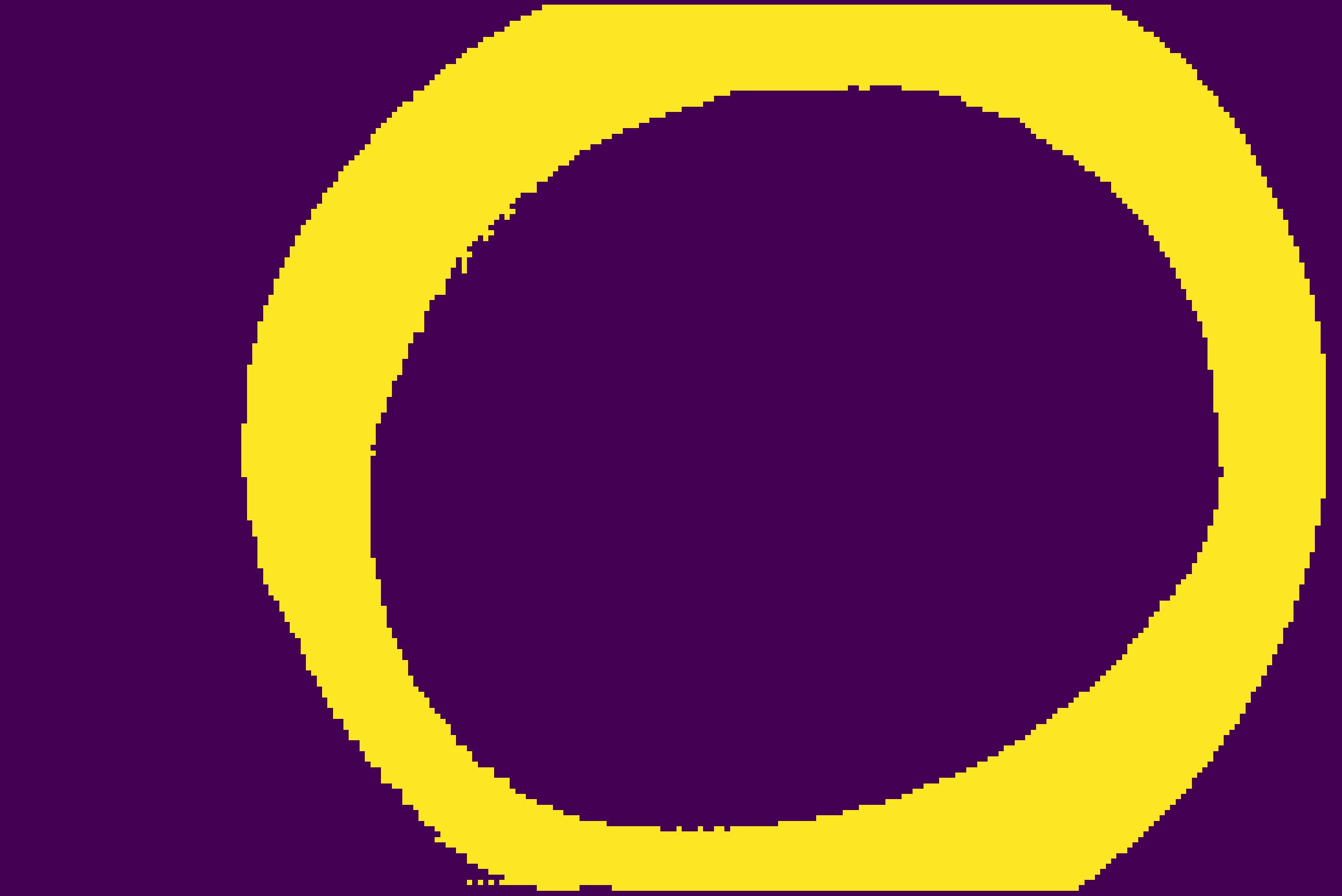} }}%
        \subfloat[]{{\includegraphics[width=.24\linewidth]{ 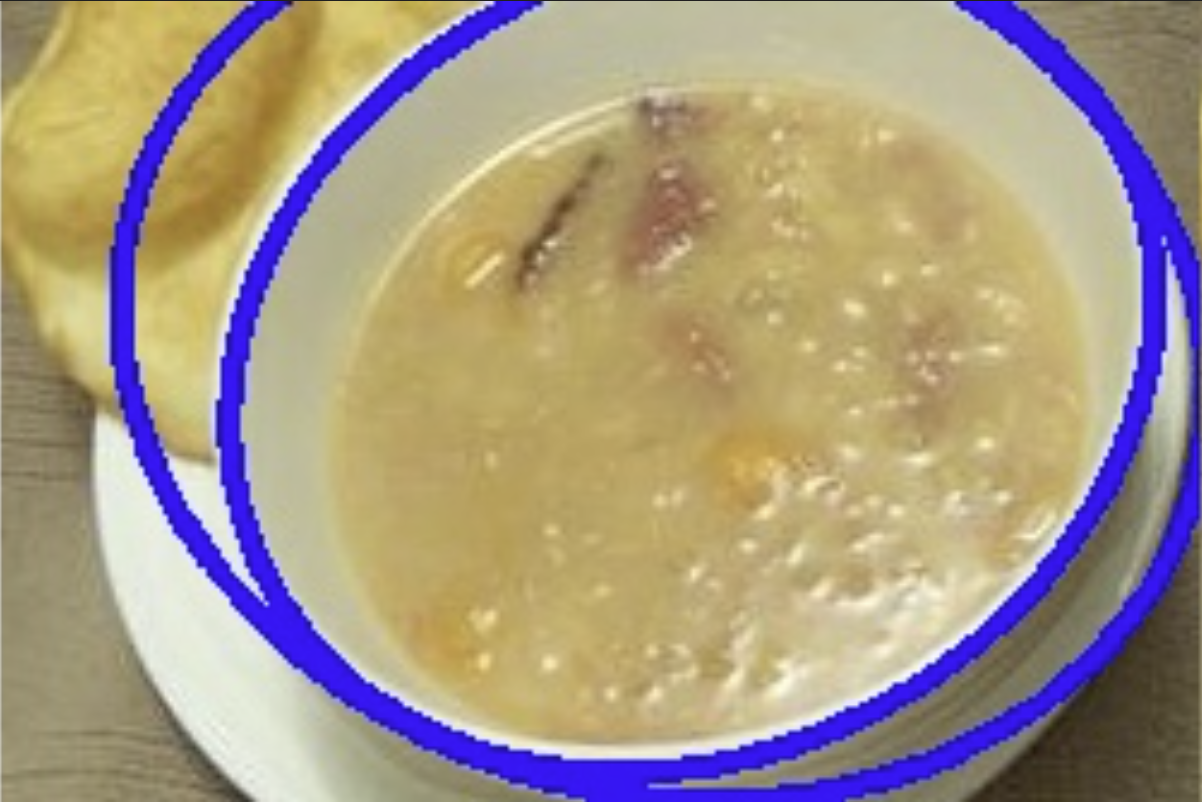} }}%
        \subfloat[]{{\includegraphics[width=.24\linewidth]{ 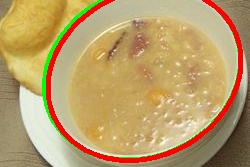} }}%

    \caption{(a): the input image containing soup in a bowl. (b): the bowl segmented using GroundedSAM \cite{GroundedSAM}, where the elliptical rim has been lost. (c): candidate ellipses generated by our method. (d): final prediction (red) and the ground truth ellipse (green).}
    \label{fig:abstract}
\end{figure}

\section{Introduction}



In the realm of food image processing, accurately estimating ellipses has posed a persistent challenge, drawing attention for a decade \cite{wenyan_jia_food_2009, EstimatingFoodInCircularBowl}. Food items are commonly presented in plates, bowls, or cups, and when viewed from various angles, the rims of these containers appear as ellipses. Leveraging this elliptical representation, we can estimate both the camera perspective and the portion size relative to the original container \cite{EstimatingFoodInCircularBowl, diningbowlreconstruction, imageBasedVolumeEstimation}. 
This characteristic renders ellipse estimation an ideal technique for two important downstream applications. Firstly, estimated ellipses serve as reference points for quantifying the contents of food items, aiding in the assessment of dietary intake. Secondly, they facilitate the estimation of rich multimodal information essential for finely controlled generation of food images or videos, such as camera angles or reference portion sizes. The success of both applications requires good method generalization to in-the-wild data which is characterized by complex food types, shapes and arrangements alongside plates, bowls, and utensils.

Several ellipse fitting methods exist such as the Direct Least Squares method \cite{DLS}, posing the ellipse fitting problem as an unconstrained optimization problem. Current general-purpose state-of-the-art method for ellipse fitting in-the-wild is that of Meng \etal, AAMED \cite{meng_arc_2020}, leveraging edge detection followed by elliptic arc segmentation and a subsequent search of an arc adjacency matrix.
Both traditional and data-driven methods for ellipse fitting lack a profound understanding of the semantic context of objects within the image \cite{meng_arc_2020,li_shape-biased_2022}. Although segmentation-based foundational models \cite{kirillov_segment_2023} have excelled in general object segmentation tasks, they encounter difficulties in accurately delineating plate and bowl segmentation masks with precise boundaries for ellipse estimation.



In this paper, we propose WildEllipseFit, a framework to automatically detect and estimate the elliptical parameters describing the rims of circular bowls and dining plates in an image. The framework combines Canny edge detection \cite{CANNY} and contour processing along with object detection from GroundingDINO \cite{liu2023grounding} to create accurate ellipse predictions. We also introduce an innovative contour filtering procedure, yielding good results on in-the-wild plate ellipse fitting.

To facilitate evaluation of our method we also introduce the Yummly-ellipse dataset. We have annotated images from the Yummly dataset \cite{YUMMLYDS_PAPER} with manually-drawn ellipses around the rims of plates and bowls in the scene. The Yummly-ellipse dataset constitutes our second contribution towards more accurate plate ellipse fitting making it a good candidate for a benchmark dataset for the plate-ellipse-fitting problem.
\label{sec:intro}

\label{sec:related}

\newcommand{\setbuilder}[3]{#1 \leftarrow \left \{ #2 \: : \: #3 \right \}}
\section{Method}
\begin{figure}[t]
    \centering
    \includegraphics[width=\linewidth]{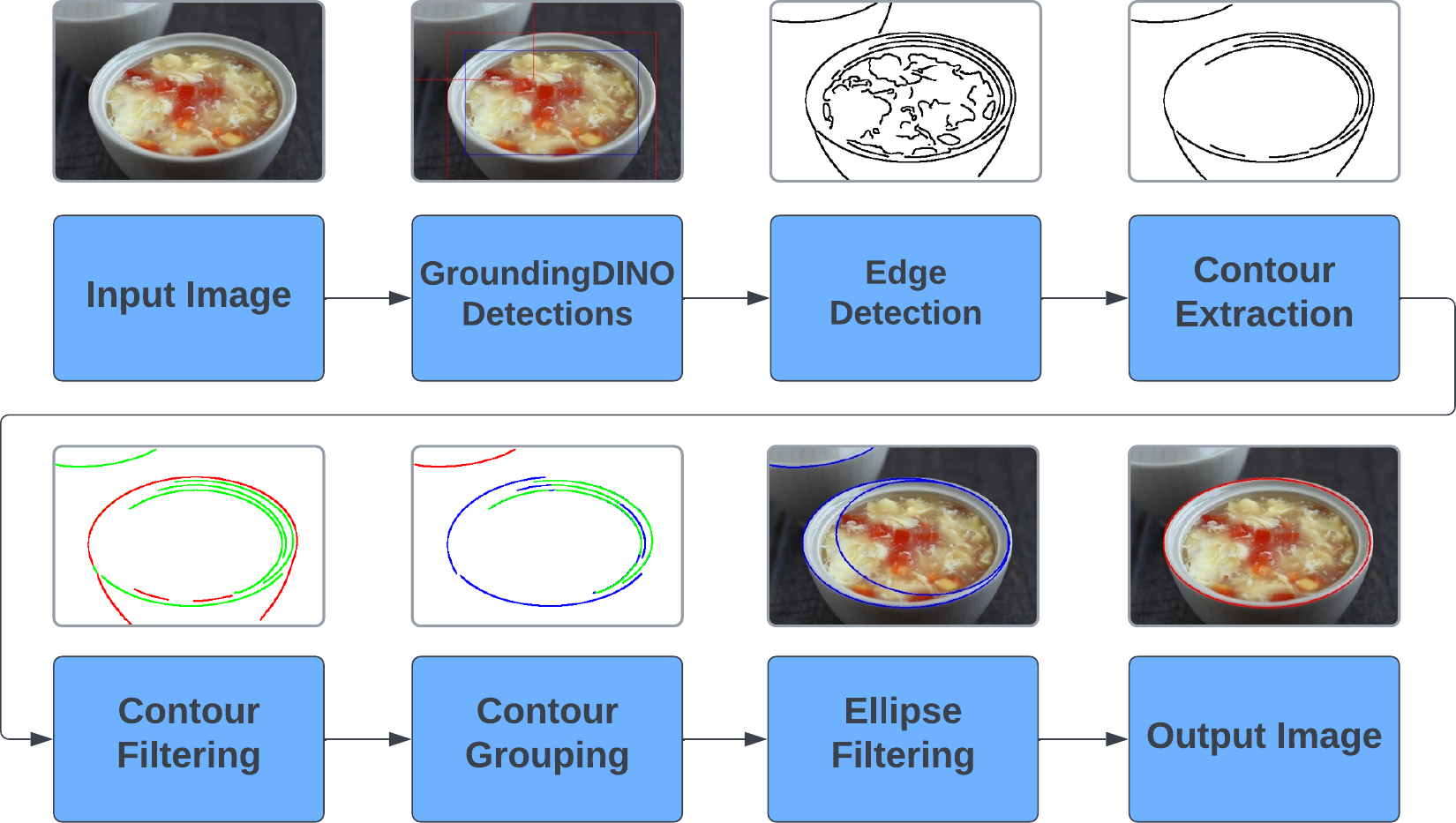}
    \caption{A block diagram for the ellipse fitting process with an example image.}
    \label{fig:entireprocess}
\end{figure}

\subsection{Method Overview}
The overall process is visualized in Figure \ref{fig:entireprocess}. First, GroundingDINO \cite{liu2023grounding} is used to obtain bounding box detections for food and plates in the image. Then Canny edge detection \cite{CANNY} is applied to the image resulting in an edge map. Subsequently, curved contours that could belong to an ellipse are extracted and then filtered to remove contours not belonging to the rim. After all irrelevant contours have been filtered, the contours are grouped to determine which belong to the same ellipse. Once all contours have been grouped, ellipse fitting is applied to each group of contours to generate ellipse predictions. These prediction ellipses are finally filtered using semantic information from GroundingDINO.

\subsection{GroundingDINO Detections \& Edge Detection}
The process used to localize ellipses leverages both GroundingDINO \cite{liu2023grounding} and contour processing to determine whether ellipse predictions are correct. GroundingDINO is an open-set object detector capable of detecting arbitrary objects. For each image $I$, it is used to obtain bounding boxes around plates, bowls, and food. Food bounding boxes are combined into a larger bounding box under two conditions: 1) when they are contained within the bounding box of the same plate/bowl or, 2) when they meet an Intersection-over-Union (IoU) threshold, $G_{M}$. This is done to cover the entirety of the food items on a given dish with one collective bounding box. Subsequently, the image is processed using the Canny edge detection \cite{CANNY} algorithm to obtain an edge map $\mathcal P$. What follows is extraction of elliptical edges.

\subsection{Contour Extraction}
Contours from the edge map $\mathcal P$ are obtained, with each contour $P$ being a sequence of pixels:

\begin{equation}
P = \left(\left( x, y\right)\right)_{x\in[0; w], y \in [0; h]}
\label{eq:1}
\end{equation}

Where $h$ is the height of the image $I$ and $w$ is its width. To separate elliptical edges of plates and bowls from jagged edges of food, information about their curvature can be used. We employ a similar method to that of Prasad \cite{prasadImageProc}.

Elliptical edges within the image will not have large changes in the tangential slope when compared to the jagged edges in the image. If each contour $P$ was represented as a non-discrete curve, $f(x, y)$, the second derivative of $f(x, y)$ $\frac{d^2 f}{d x^2 d y^2}$ should be low for elliptical contours. Since points in the sequence $P$ represent discrete pixels, the smallest possible difference in either $x$ or $y$ is $1$, which means the concept of a derivative cannot directly be applied. We use vectors to represent the change in $x$ and the change in $y$ between two points in the sequence. Let $\mathbf d_{p_u, p_v}$ be the vector representing the change in $x$ and $y$ between two pixels, where $p_u, p_v \in P$ and $p_u, p_v$ are separated in the sequence by $s$ steps, i.e. $v = u + s$. $s$ is a hyperparameter.

Hence:
\begin{subequations}\label{2}
\begin{align}
\mathbf d_{p_u, p_v} &= \mathbf p_v - \mathbf p_u \label{eq:2a} \\
&= \begin{bmatrix}
x_v\\ y_v
\end{bmatrix} - \begin{bmatrix}
x_u \\ 
y_u
\end{bmatrix} \label{eq:2b}
\end{align}
\end{subequations}




Subsequently, we consider the L1 distance, $a$, between consecutive $\mathbf d$ vectors:
If $a$ is less than or equal to a set threshold value, $\epsilon$, then a candidate contour $C = \emptyset$ is updated to the union of $C$ and the subsequence $[ p_{v}; p_{v+s} ]$ from $P$:

\begin{equation}
\setbuilder{C}{C \cup [ p_{v}; p_{v+s} ]}{a \le \epsilon}
\label{eq:3}
\end{equation}

If the $a$ is greater than $\epsilon$, then the points in the aforementioned interval of the sequence $P$ are discarded. The current set $C$ of accepted points is checked to see if its length is at least $L$. If so it is recorded as a new contour:

\begin{equation}
\setbuilder{\mathfrak C }{\mathfrak C \cup C}{|C| \ge L}
\label{eq:4}
\end{equation}

The set of accepted points $C$ is then reassigned as $C = \emptyset$. This is to avoid any gaps or duplicate points in the subsequently captured contours. This process is repeated for every edge $P \in \mathcal P$. Ultimately, a set of contours $\mathfrak C$ is obtained representing only the edges with low curvature. 
\begin{figure}[h]
    \centering
    \includegraphics[width=0.32\linewidth]{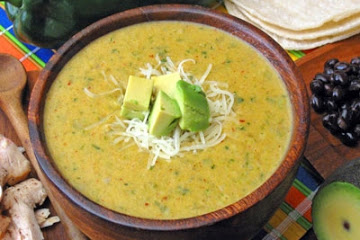}
    \includegraphics[width=0.32\linewidth]{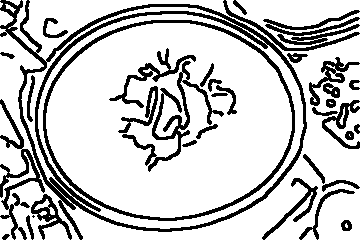}
    \includegraphics[width=0.32\linewidth]{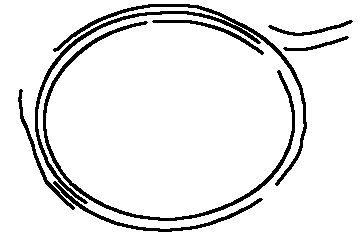}
    \caption{Example of curve extraction process, top left is original image. Middle is the Canny edge detection algorithm applied to the entire image. The right image is the contours extracted as a result of applying the curve extraction process.}
    \label{fig:extraction}
\end{figure}

\subsection{Contour Filtering}
As a result of the contour extraction algorithm, straight contours are also preserved. Thus, the next step is to remove them.
Two extremal points of each contour $C \in \mathfrak C$ are identified and a line section (chord) is drawn between the two points. The chord can be represented as a vector $\pmb \chi$.

The smallest possible distance between each point in the contour $p \in C$ and $\pmb \chi$ is then recorded. This is equivalent to taking the Euclidean distance between $\mathbf p$ and $\pmb \chi$. Then we remove points from $\mathfrak C$ based on the following condition:

\begin{equation}
\setbuilder{\mathfrak C}{\mathfrak C \setminus C }{\max\limits_{\mathbf p \in C} \lVert \mathbf p - \pmb \chi \rVert^2_2 < d}
\label{eq:5}
\end{equation}

This process is repeated $\forall C \in \mathfrak C$.


The final step in Contour Filtering is used to focus only on the top-surface elliptical contours of plates and bowls. When viewed from the side, bowls often have two elliptical contours, the one representing the top surface delineated by the bowl rim and one that can be observed as a result of viewing it from the side, stemming from the natural curvature of a bowl dish. Thus, it becomes necessary to distinguish which contour describes the rim of the bowl.

For every food bounding box predicted by GroundingDINO and its closest plate/bowl bounding box, there should be a gap between the bottom edges of the bounding boxes. Only contours and their points within this gap are considered. The y-distance between the bottom edge of the food bounding box and the lowest point in the Y coordinate for each contour is calculated and added to the set $\mathcal D$:

\begin{equation}
\setbuilder{\mathcal D}{\lvert y_{\text{bbox,f}} - y_{min,C} \rvert}{\forall C \in \mathfrak C}
\label{eq:6}
\end{equation}

Then the set of contours $\mathfrak C$ and set of y-distances $\mathcal D$ are updated iteratively:

\begin{align}
& \textbf{while } \lvert \max \mathcal D - \min \mathcal D \rvert > H \textbf{ do} \nonumber \\
& \quad \setbuilder{U}{\left \lvert d - \frac{1}{\lvert \mathcal D \rvert}\sum\limits_{i=1}^{\lvert \mathcal D \rvert } d_i \right \rvert }{\forall d \in \mathcal D } \nonumber \\
& \quad \setbuilder{\mathfrak C}{\mathfrak C \setminus C_j}{j = \arg \max U} \nonumber \\
& \quad \setbuilder{\mathcal D}{\mathcal D \setminus d_k}{k = \arg \max U} \nonumber \\
& \textbf{done}
\label{eq:7}
\end{align}

Iteratively discarding contours of which y-distance to the bottom of the food instance bounding box deviates the most from the mean of all such distances allows us to filter out outlier contours. If $\lvert \mathcal D \rvert = 2$ and $\lvert \max \mathcal D - \min \mathcal D \rvert$ is still $>H$, we reject the contour for which $d \in \mathcal D$ is the largest.

This method is more robust than just removing the lowest-positioned contours because for some samples the curve extraction process does not remove some of the food instance edges. Rejecting the lowest-positioned contours would inevitably eliminate all valid bowl contours under such circumstances.

\subsection{Contour Grouping}
Now, having obtained the individual contours needed, the contours of the same ellipse must be combined. To check if two contours, $C_1$ and $C_2$, are to be combined, we combine their sets of points into one set, $C_{C_1 \cup C_2} = C_1 \cup C_2$. If fitting an ellipse to C results in a close fit to the points, then they are likely to be a part of the same ellipse. For every point in the combined set, we get the squared Euclidean distance to the closest point on the fitted ellipse $E$. The mean of these distances is then taken as a score $\mathcal S$ for ellipse fitting:

\begin{equation}
\mathcal S = \frac{1}{|C_{C_1 \cup C_2}|} \sum\limits_{\mathbf p \in C_{C_1 \cup C_2}} \lVert \mathbf p - E \rVert^2_2\\
\label{eq:8}
\end{equation}

If the score is lower than a set limit, \emph{M}, the two contours are combined and their union replaces the two contours in the set $\mathfrak C$:

\begin{equation}
\setbuilder{\mathfrak C}{(\mathfrak C \setminus \{C_1, C_2\}) \cup \{C_1 \cup C_2\}}{\mathcal S < M}
\label{eq:9}
\end{equation}


\subsection{Ellipse Filtering}
After the contour merging algorithm, a convex hull of each contour is first found and then fitted with OpenCV's default ellipse fitting implementation \cite{opencv_library, OPENCV_ELLIPSE_METHOD}, which we further denote as operator name: "$\operatorname{fit}$". We leverage the bounding box information from GroundingDINO to filter out incorrect ellipses. Each ellipse is first compared with the closest plate/bowl bounding box. Intuitively, a well-fitted ellipse prediction should not have a large portion of its area outside of the bounding box. This idea can be used to filter out ellipses that have a minimum area fraction, \emph{$A_{p}$}, outside the bowl or plate bounding box. A good ellipse prediction would also have its center coordinates, $(x_{e}, y_{e})$, close to the center of any of the food bounding boxes. Therefore, the distance between the center of each ellipse and the closest food bounding box can also be thresholded by a maximum distance, \emph{$D_{f}$}:

\begin{subequations}\label{eq:10}
\begin{align}
&\setbuilder{\mathcal E}{\mathcal E \setminus \{E\}}{\frac{\operatorname{Area}(E \setminus \text{bbox\_p})}{\operatorname{Area}(\text{E})} \ge A_p}
\label{eq:10a}\\
&\setbuilder{\mathcal E}{\mathcal E \setminus \{E\}}{\lVert \mathbf e_{c} - \mathbf b_c \rVert^2_2  \ge D_f}
\label{eq:10b}
\end{align}
\end{subequations}

Where $\text{bbox\_p}$ stands for the plate bounding box, $\mathbf e_c = \begin{bmatrix} x_e & y_e \end{bmatrix}^T$ and $\mathbf b_c = \begin{bmatrix} x_{\text{bbox\_f center}} & y_{\text{bbox\_f center}}\end{bmatrix}^T$, where $\text{bbox\_f}$ is the food bounding box.

\begin{figure}[h]
    \centering
    \includegraphics[width=0.24\linewidth]{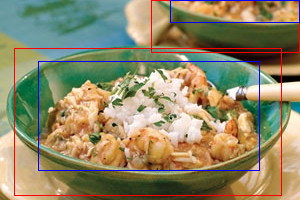}
    \includegraphics[width=0.24\linewidth]{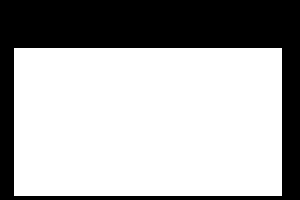}
    \includegraphics[width=0.24\linewidth]{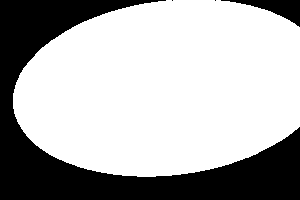}
    \includegraphics[width=0.24\linewidth]{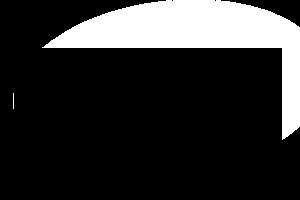}
    \caption{Plate filtering.}
    \label{fig:platefiltering}
\end{figure}

\begin{figure}[h]
    \centering
    \includegraphics[width=0.24\linewidth]{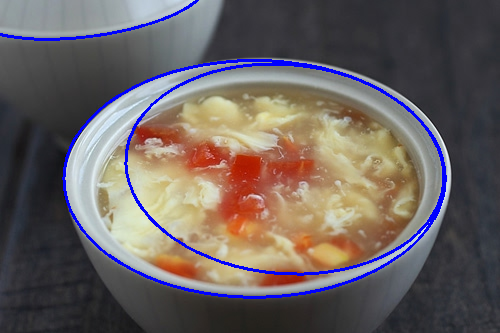}
    \includegraphics[width=0.24\linewidth]{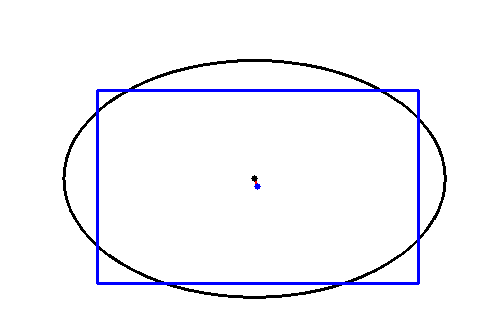}
    \includegraphics[width=0.24\linewidth]{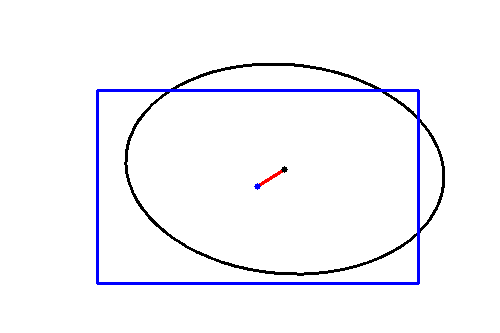}
    \includegraphics[width=0.24\linewidth]{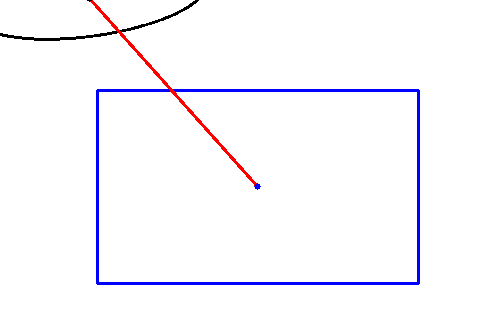}
    \caption{Food distance filtering.}
    \label{fig:foodfiltering}
\end{figure}

After these two filters have been applied, the remaining ellipse prediction parameters are returned.

\label{sec:method}

\begin{figure}[t]
    \centering
    \includegraphics[width=0.24\linewidth]{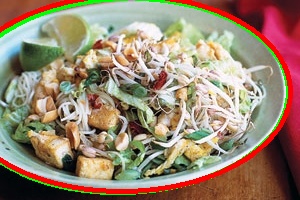}
    \includegraphics[width=0.24\linewidth]{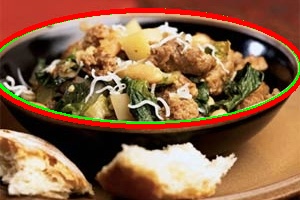}
    \includegraphics[width=0.24\linewidth]{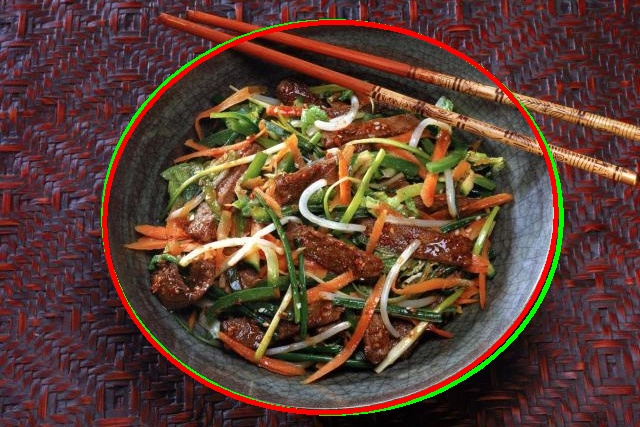}
    \includegraphics[width=0.24\linewidth]{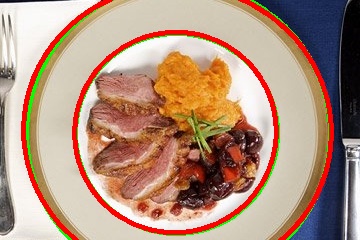}

    \caption{Sample correct detections by our framework.}
    \label{fig:correctdetections}
\end{figure}
\begin{figure}[t]
    \centering
    \includegraphics[width=0.24\linewidth]{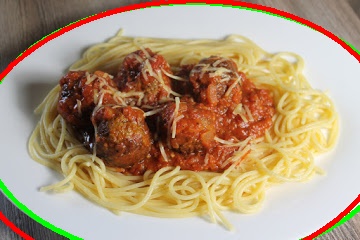}
    \includegraphics[width=0.24\linewidth]{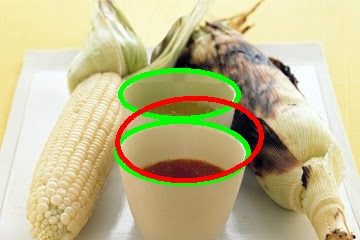}
    \includegraphics[width=0.24\linewidth]{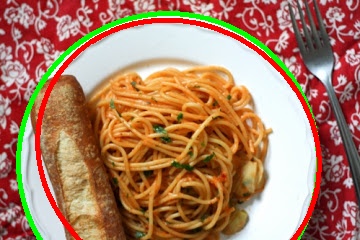}
    \includegraphics[width=0.24\linewidth]{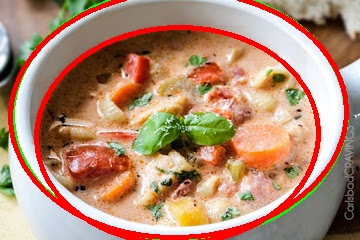}

    \caption{Sample incorrect detections by our framework.}
    \label{fig:misdetections}
\end{figure}
\section{Results}
\label{sec:results}
We applied our algorithm to a collection of 400 images from the Yummly-28K dataset \cite{YUMMLYDS_PAPER}, which contains a wide variety of food scenes under many different viewing angles.
All 400 images contained at least one elliptical contour and were manually annotated with ground truth ellipses. We name the resulting dataset Yummly-ellipse. 

\begin{table}
  \centering
  \begin{tabular}{@{}c|ccccccccc@{}}
    \toprule
     &$G_M$ &  $s$ &  $\epsilon$ & $L$ & $d$ & $H$ & $M$ & $A_p$ & $D_f$ \\
    \midrule
     A & 0.6 & 8 & 2 & 60 & 7.0 & 10 & 150 & 0.08 & 450 \\
     B & 0.7 & 7 & 2 & 60 & 7.5 & 10 & 125 & 0.06 & 450 \\
    \bottomrule
  \end{tabular}
  \caption{Optimal Parameter sets for both method A and B calculated on training dataset of 300 images.}
  \label{tab:optimal-params-a}
\end{table}


We have chosen to use the Chamfer distance between the perimeters of the predicted ellipse and the nearest ground truth ellipse as the metric for evaluation as it incorporates all errors of the ellipse parameters under one metric. The Chamfer distance between two sets of points, $A$ and $B$, is defined as:

\begin{equation}
D_{\mathrm{Chamfer}}(A, B) = \frac{1}{2} \left( \sum\limits_{x \in A} \min\limits_{y \in B} d(x, y) + \sum\limits_{y \in B} \min\limits_{x \in A} d(x, y) \right)
\label{eq:11}
\end{equation}

In our case we adopt Euclidean distance for $d(x, y)$. We choose two methods of comparing ground truth ellipses to predictions. The first method, method A, takes each ground truth ellipse and calculates the Chamfer distance between its perimeter and the nearest predicted ellipse's perimeter. The second method, method B, takes each predicted ellipse and calculates the Chamfer distance between its perimeter and the nearest ground truth ellipse's perimeter. Method A penalizes missed ellipse predictions, assigning more importance to recall. Method B focuses more on the precision of each individual ellipse prediction, which is an important consideration if WildEllipseFit is to be used for camera angle estimation.

The threshold for detection used by GroundingDINO for the plates, bowls, and food was set at $0.35$ with prompts of "Plate", "Bowl", and "Food". For Canny Edge detection, we used Scikit-Image's \cite{scikit-image} Canny edge detection implementation with a $\sigma$ of $2.5$.

\begin{table}[htbp]
  \centering
  \begin{tabular}{@{}l|ccc|ccc@{}}
    \toprule
    Method & $\mu_{A}$ & $\sigma_{A}$ & $N$ & $\mu_{B}$ & $\sigma_{B}$ & $N$\\
    \midrule
    AAMED & 19.33 & 32.57 & 68 & 38.91 & 56.05 & 68\\
    Ours & \textbf{16.87} & 33.36 & 37 & \textbf{7.11} & 7.61 & 35\\
    \bottomrule
  \end{tabular}
  \caption{Baseline Chamfer Distance metric for the method proposed by Meng \etal \cite{meng_arc_2020} versus WildEllipseFit on a test dataset of 100 images. $\mu_A$ indicates Chamfer distance calculated according to Method A, $\mu_B$ according to Method B. $\sigma$ indicates standard deviation. $N$ refers to number of images with atleast one ellipse prediction. Optimal mean Chamfer Distance has been bolded.}
  \label{tab:baselines}
\end{table}



We compare our method in Table \ref{tab:baselines} using the optimal parameters found previously with the arc adjacency matrix method proposed by Meng \etal \cite{meng_arc_2020} using optimal hyperparameters reported by the authors. We find that while our method results in fewer images with ellipse predictions, we obtain a lower Chamfer distance indicating a higher precision.

\section{Conclusion}
In this paper, we discussed the motivation for estimating elliptical parameters of circular dining plates and bowls. We outlined WildEllipseFit, a multi-stage process combining contour processing with semantic information from zero-shot models to estimate ellipse parameters. We demonstrate its performance on a dataset of 400 images annotated with ground truth ellipses. The proposed framework was compared with current ellipse-fitting methods and the improvements provided by incorporating zero-shot models was shown. 
\label{sec:conclusion}

\FloatBarrier
{\small
\bibliographystyle{ieeenat_fullname}
\bibliography{11_references}
}


\end{document}